\documentclass[runningheads]{llncs}
\usepackage{graphicx}

\usepackage{microtype}
\usepackage{multirow}
\usepackage{booktabs}
\usepackage{tikz}
\usepackage{comment}
\usepackage{amsmath,amssymb} % define this before the line numbering.
\usepackage{color}
\usepackage{hyperref}
\hypersetup{colorlinks,linkcolor={blue},citecolor={blue},urlcolor={red}} 
\begin{document}
% \linenumbers
\pagestyle{headings}
\mainmatter
\def\ECCVSubNumber{100}  % Insert your submission number here

\title{Simple Unsupervised Multi-Object Tracking} % Replace with your title

\begin{comment}
\titlerunning{ECCV-20 submission ID \ECCVSubNumber} 
\authorrunning{ECCV-20 submission ID \ECCVSubNumber} 
\author{Anonymous ECCV submission}
\institute{Paper ID \ECCVSubNumber}
\end{comment}
%******************

%\begin{comment}
\titlerunning{Simple Unsupervised Multi-Object Tracking}
\author{Shyamgopal Karthik\inst{1} \qquad
Ameya Prabhu\inst{2} \qquad
Vineet Gandhi\inst{1}}
\authorrunning{S. Karthik et al.}
\institute{Center for Visual Information Technology\\ Kohli Center on Intelligent Systems, IIIT Hyderabad, India  \and
University of Oxford\\
\email{\{shyamgopal.karthik@research,vgandhi@\}.iiit.ac.in}\\
\email{ameya@robots.ox.ac.uk} }
%\end{comment}
%******************
\maketitle

\begin{abstract}
Multi-object tracking has seen a lot of progress recently, albeit with substantial annotation costs for developing better and larger labeled datasets. In this work, we remove the need for annotated datasets by proposing an unsupervised re-identification network, thus sidestepping the labeling costs entirely, required for training. Given unlabeled videos, our proposed method (SimpleReID) first generates tracking labels using SORT~\cite{bewley2016simple} and trains a ReID network to predict the generated labels using crossentropy loss. We demonstrate that SimpleReID performs substantially better than simpler alternatives, and we recover the full performance of its supervised counterpart consistently across diverse tracking frameworks.  The observations are unusual because unsupervised ReID is not expected to excel in crowded scenarios with occlusions, and drastic viewpoint changes. By incorporating our unsupervised SimpleReID with CenterTrack trained on augmented still images, we establish a new state-of-the-art performance on popular datasets like MOT16/17 without using tracking supervision, beating current best (CenterTrack) by 0.2-0.3 MOTA and 4.4-4.8 IDF1 scores. We further provide evidence for limited scope for improvement in IDF1 scores beyond our unsupervised ReID in the studied settings. Our investigation suggests reconsideration towards more sophisticated, supervised, end-to-end trackers~\cite{xu2020train,braso2019learning} by showing promise in simpler unsupervised alternatives.
\keywords{Multi-Object Tracking, Re-Identification, Unsupervised Learning}
\end{abstract}

\section{Introduction}
\label{sec:intro}

%% Para 1: Why is MOT important? What has happened in recent years, and challenges they face.
Understanding human interactions and behaviour over videos has been a fundamental problem in computer vision with applications in action recognition, sports video analytics, and assistive tech and requires tracking multiple people over time. Multi-object trackers broadly consist of two key components: (i) A spatio-temporal association model which associates boxes in nearby frames to create clusters of tracklets, and (ii) A re-identification model which associates tracklets over larger windows to deal with complexities in tracking such as occlusions and target interactions. Re-identification is a major challenge in tracking, with sophisticated supervised approaches requiring expensive annotations to assign trajectories across frames to every single person in a video. Availability of labeled datasets\cite{manen2017pathtrack,milan2016mot16} has alleviated the problem.  For instance IDF1 (MOTA) scores have improved from 51.3(48.8)~\cite{tang2017multiple} to 59.9 (55.9)~\cite{braso2019learning} on  the MOT16~\cite{milan2016mot16} benchmark in the past 3 years.

%% Para 2: How they've approached the challenge, and what are shortcomings of that approach -- we're good here.
There has been a growing need to annotate larger tracking datasets with the aim of improving re-identification (ReID) models. However, annotating tracking datasets require hefty labeling costs and scale poorly with dataset size. To illustrate the effort and cost required, annotating 6 minutes worth of video of the MOT15 benchmark~\cite{leal2015motchallenge} using the standard annotation procedures would take at least 22 hours of annotation time~\cite{manen2017pathtrack}. Annotating just twenty-six hours of video data (VIRAT dataset~\cite{oh2011large}) with state-of-the-art protocols in place \cite{oh2011large,vondrick2011video} costs tens of thousands of dollars.
%% Para 3: The issue we propose to alleviate this issue and how we alleviate this issue.
We propose to learn our model in an unsupervised manner in the free-labels paradigm (Section 6.3.2 in \cite{jing2019self}) in a two-step manner. We first generate tracking labels given unlabeled videos and the corresponding set of detections. Then, we learn a ReID network to predict the generated label given an input image. To the best of our knowledge, ours is the first work to propose unsupervised ReID models for multi-object tracking and completely do away with the tremendous annotation costs for tracking datasets. Throughout the paper, we consider supervision only in the context of sidestepping trajectory-level annotations. Using off-the-shelf detectors~\cite{ren2015faster,redmon2016you,chen2019hybrid} trained on COCO is not viewed as supervision in our context. The proposed ReID network complements the unsupervised spatio-temporal association models~\cite{wojke2017simple,bergmann2019tracking} proposed in the prior art, leading to a more complete unsupervised tracking framework.

We go one step further and aim to test the limits of our unsupervised tracking paradigm. We empirically test for two desiderata w.r.t IDF1 scores: (i) Our unsupervised ReID should perform significantly better than naive ReID methods when incorporated into any tracker; (ii) Our unsupervised ReID should achieve performance equivalent to the original supervised counterpart. We demonstrate that we are able to achieve these desiderata consistently across datasets, detectors, and diverse trackers. The resultant unsupervised tracker, when combined with CenterTrack~\cite{zhou2020tracking} trained on single images, achieves state-of-the-art performance on the MOT16/17 test challenge server. We beat the latest supervised trackers by large margins, outperforming CenterTrack by 0.3 MOTA, and 4.8 IDF1 scores. We then demonstrate that there is limited scope for further improvement beyond our proposed unsupervised ReID by demonstrating that the Oracle counterpart of our ReID model makes only minor gains.

% Para 5: Implications of our work
We would also like to highlight that while our work is conceptually simple, the contributions made are significant. We expect our investigation to be of significant interest to the MOT community by demonstrating that simple unsupervised ReID is sufficient even in crowded scenarios with occlusions and person interactions. Our investigation contrasts the current shift towards using more supervised, end-to-end trackers for MOT Challenge datasets. We hope our work spurs research in the unsupervised MOT paradigm, exploring extensions to other tracking scenarios (3D/vehicles/pose tracking) and do away with the labeling effort wherever unnecessary.

\section{Related Works}
\label{sec:related}
Monocular 2D multi-object tracking on videos is an extensively studied problem. \cite{ciaparrone2020deep} offers a comprehensive review of works on MOT Challenge datasets. A  popular paradigm is to model the detections as a graph. Various approaches have been proposed here including using network flows~\cite{zhang2008global}, graph cuts~\cite{tang2017multiple}, MCMC~\cite{yu2007multiple} and minimum cliques~\cite{zamir2012gmcp} if the entire video is provided beforehand (batch processing). In scenarios where we get frame-by-frame input, Hungarian matching~\cite{wojke2017simple,bewley2016simple}, greedy matching~\cite{zhou2020tracking} and Recurrent Neural Networks~\cite{fang2018recurrent,sadeghian2017tracking} are popularly used models for sequential prediction (online processing). The association metrics/cost functions used by these consist of (i) Spatio-temporal relations (ii) Re-identification. 

\textbf{Spatiotemporal relations:} There has been much investigation into appearance-free methods for the spatio-temporal association. Basic methods proposed include using Intersection-Over-Union (IoU) between detections~\cite{bochinski2017high} or incorporating a velocity model using a Kalman filter~\cite{bewley2016simple}. The velocity model can also be learned using Recurrent Neural Networks ~\cite{fang2018recurrent,sadeghian2017tracking}. The complexity of assigning pairwise costs can be further increased by incorporating additional cues from head/joint detectors~\cite{chari2015pairwise,henschel2018fusion}, segmentation~\cite{milan2015joint}, activity recognition~\cite{choi2012unified}, or keypoint trajectories~\cite{choi2015near}. Recent approaches leverage appearance-reliant pre-trained bounding box regressors from object detection~\cite{bergmann2019tracking}  or single object tracking~\cite{xu2020train,chu2019online} pipelines to regress the bounding box in the next frame. Since most of the above models are unsupervised (requiring no tracking annotations), they complement our work and can be incorporated with our proposed approach for creating efficacious unsupervised trackers.

\textbf{ReID across multiple cameras:}   Supervised training of CNNs~\cite{zhou2019omni} on large labeled datasets~\cite{zheng2015scalable,li2014deepreid} has given excellent results for ReID across multiple cameras. In addition to this, there have been approaches to exploit the pose information using off-the-shelf body pose detectors~\cite{su2017pose,suh2018part}. Attention mechanisms have also been explored to capture the important regions in the foreground~\cite{si2018dual,song2018mask}. Generative models have been employed to augment the training data for improved performance~\cite{zheng2019joint,li2018adversarial}. We recommend this excellent survey~\cite{ye2020deep} for a complete review.
In contrast, we work on tracking with a single camera, with reasonable frame-rates (no drastic appearance variations). Additionally, the objective is to distinguish the target pedestrian among a small set of different looking pedestrians in a given frame, with the aid of additional detection information. Hence, we believe our simple, noisy unsupervised re-identification model might suffice. Sophisticated unsupervised ReID networks ~\cite{lin2019bottom,li2018unsupervised} designed for multiple cameras ReID may not be required for MOT.

\textbf{ReID for monocular 2D tracking:} Re-identification has been a major challenge in tracking, with matching using similarity between CNN features being the dominant approach~\cite{ristani2018features}. Past works have proposed different methods to train the CNN ranging from using siamese networks~\cite{leal2016learning} with triplet loss, further augmented by hard negative mining~\cite{bergmann2019tracking} or other metric learning losses like cosine loss~\cite{wojke2017simple}. Incorporating a combination of loss functions~\cite{ma2018customized} or pose information~\cite{tang2017multiple} as well as fine-tuning the ReID model on the test sequence~\cite{ma2018customized}.  All the above ReID networks are supervised and fairly complex to train. We are the first work to demonstrate that simple unsupervised ReID networks are sufficient for this context. It is important to note that in most MOT pipelines, this is the only component that uses tracking annotations.

\textbf{Evaluation metric for MOT:} Multi-Object Tracking Accuracy (MOTA) is not a good metric to illustrate ReID performance because it focuses on object coverage and therefore is dominated by false negatives. An excellent detector can achieve high MOTA scores despite being a poor tracker with a large number of ID switches \cite{zhou2020tracking}. Identity-F1 (IDF1) has been shown to measure long consistent tracks without switches and widely shown~\cite{maksai2019eliminating,ciaparrone2020deep} to be a better metric for tracking performance.  We accordingly focus and emphasize on IDF1 scores.

\textbf{End-to-end supervised MOT:} Recent works circumvent the above paradigm either partially or completely by learning the MOT solver using end-to-end supervision. Early works~\cite{wang2017learning,schulter2017deep} performed end-to-end learning in the min-cost flow data association framework. Recently, approaches like \cite{xu2020train} and \cite{braso2019learning} perform end-to-end optimization by introducing differentiable forms of Hungarian matching and clustering formulation, respectively. Parallel works~\cite{zhou2020tracking,zhang2020simple,wang2019towards} attempt to perform simultaneous object detection, data association, and sometimes re-identification in a single network. Most notable among these, CenterTrack~\cite{zhou2020tracking} is capable of training the detector using only augmentations of still images.  These methods involving joint detection and tracking deliver high performance at real-time inference speeds but require high annotation costs. Our work differs in principle by removing and replacing supervised components yet outperforming these trackers, without incurring the associated labelling cost.

\section{Approach}
\label{sec:approach}

Our goal is to leverage the abundance of unlabeled videos to learn ReID models (without manual cost). Our unsupervised learning method can be categorized as learning by generating labels (Ref. Section 6.3.2 of \cite{jing2019self}). In a nutshell, given unlabeled videos and corresponding bounding boxes, we first generate tracking labels. We then learn a ReID network by predicting the generated label given a detection.

\subsection{Framework: Learning by generating tracking labels}
\label{sec:framework}
Here, we describe the two parts of our proposed framework in detail: (i) Generating the labels, and (ii) Learning the network.
\begin{table}[t]
\vspace{-0.25cm}
\parbox{.5\linewidth}{
\begin{center}
        \resizebox{0.45\textwidth}{!}{
    \begin{tabular}{lcc} \toprule
\textbf{Model} & \textbf{Ref}  \\ \midrule
Kalman filter+Hungarian matching & \cite{bewley2016simple}  \\
IoU based tracking & \cite{bochinski2017high}  \\
Network Flow & \cite{zhang2008global}\\
Linear Programming & \cite{leal2011everybody}\\
Conditional Random Fields(CRFs) & \cite{milan2015joint}\\
Markov Decision Proceses(MDPs) & \cite{xiang2015learning}\\
Recurrent Neural Networks(RNNs) & \cite{sadeghian2017tracking}\\
Bounding Box Regression & \cite{bergmann2019tracking}  \\

\bottomrule
    \end{tabular}} \end{center}
}
\hfill
\parbox{.5\linewidth}{
\begin{center}
\resizebox{0.42\textwidth}{!}{
    \begin{tabular}{lcc} \toprule
\textbf{Training Strategy} & \textbf{Ref}  \\ \midrule
Crossentropy & \cite{tang2017multiple} \\
Triplet+hard negative mining & \cite{bergmann2019tracking} \\
Contrastive & \cite{kim2016similarity}\\
SymTriplet & \cite{zhang2016tracking}\\
%SiameseNet &  \cite{leal2016learning,tang2017multiple} \\
Cosine Loss & \cite{wojke2017simple} \\
Joint Detections & \cite{tang2017multiple} \\
Verification+Classification Loss & \cite{ma2018customized} \\

\bottomrule
    \end{tabular}} \end{center}
}
\caption{Approaches use for Spatiotemporal data association (Left). Loss functions and methods used to train CNNs for Appearance modeling (Right). We choose the simplest approach for both these components.}
\label{tab:approach}
\end{table}
\textbf{Generating labels:} Given a set of videos, each video is passed independently through an object detector. An unsupervised spatiotemporal association model from the list given in Table \ref{tab:approach} (left) is then run through the detections to obtain short contiguous tracks or tracklets (set of associated detections of the same person over time). Examples of spatiotemporal models can range from tracking using a constant velocity assumption with Kalman filtering \cite{bewley2016simple} (bounding box information only) to incorporating appearance features by using pre-trained bounding box regression from object detection pipelines to regress the bounding box \cite{bergmann2019tracking} in the next frame. Now to cluster/associate detections, we can use online methods like greedy/Hungarian matching or expensive offline methods like graph-cuts. Ultimately, the output of this step is a set of noisy track labels for each video, resulting in a pool of labeled video tracklets.

\textbf{Training ReID models:} Now, given noisy track labels per video, the task is to learn a ReID model using any of the methods given in Table \ref{tab:approach} (right). In absence of trajectory level supervision, the challenge here is to explore ways to harness the given regularities in data (in form of tracklets). There are two simple assumptions which can help the cause: (i) The videos are independent of each other (i.e., no common tracks between any two videos), and (ii) the tracklets within a video are independent of each other (i.e., each tracklet belongs to a different person). If both the assumptions are followed then each tracklet can be considered as an independent class. The simplest option which follows is to train at network to predict a label given an image, optimized with cross-entropy loss (with number of classes equalling to the number of tracklets). However, assumption (ii) may break in cases like missed detections and occlutions and may result into multiple tracklets for the same person in a video. 

An alternate option (by relaxing assumption (ii)) could be to form positive pairs from the same tracklet and negative pairs from across other videos or simultaneous tracks from the same video. Such pairing can enable learning  Siamese networks to compare two images and predict whether they are the same person or not. They can be trained with pairwise losses such as contrastive loss ~\cite{kim2016similarity} or triplet loss with hard-negative mining~\cite{bergmann2019tracking}, or more complex ones like symtriplet~\cite{zhang2016tracking} or the group loss~\cite{elezi2019group}, resulting in a trained ReID network.

\subsection{Our method}
\begin{figure}[t]
\begin{center}
\includegraphics[width=1.0\linewidth]{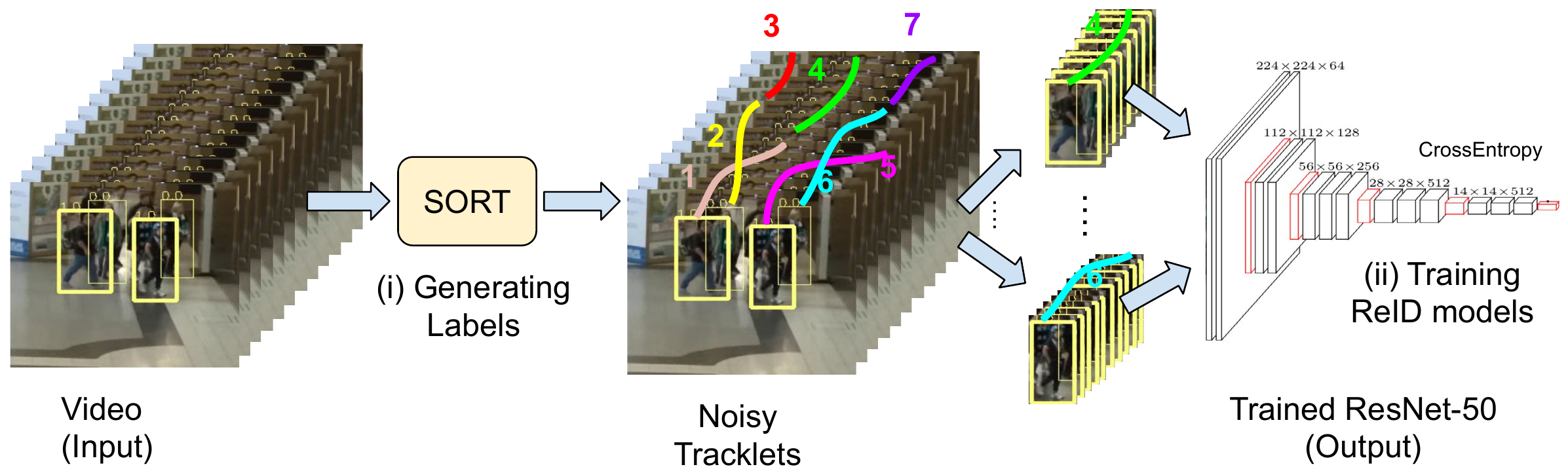}
\end{center}
\caption{Overview of our approach: Given a video with detections, we use SORT~\cite{bewley2016simple} to simulate noisy tracking labels. Then, we train the ReID network (ResNet50) to predict the track label for each input image.} 
\label{fig:comparison}
\end{figure}

We use simple methods to both simulate labels and learn the ReID network, as illustrated in Figure \ref{fig:comparison}. In step (i), we only utilize the bounding boxes and use Kalman filtering combined with Hungarian matching to simulate labels. Since we use no appearance information, our tracking labels are noisy. In step (ii), we proceed by making both the aforementioned assumptions that no two videos or tracklets share common labels. We assign a unique label to each tracklet and train a network with cross-entropy loss to predict this label given any image from that tracklet. At inference time, we integrate our ReID model into existing frameworks by simply replacing their models with ours, with no other changes. In CenterTrack, we extract tracks using its unsupervised model and refine it with our ReID network using a DeepSORT framework. Although we are aware that some enhancements can be performed to our proposed process (e.g., using a siamese framework), we show in subsequent sections that simpler choices alone are sufficient to match the performance of supervised networks.

\section{Experiments}
\label{sec:exp}

In a nutshell, in this section we incorporate our developed unsupervised ReID model (SimpleReID) into various trackers and show compelling evidence for three results: (i) our unsupervised tracker obtains state-of-the-art tracking performance on MOT16/17, outperforming recent works (ii) naive unsupervised trackers can replace their supervised counterparts consistently  (iii) there is limited scope for improvement beyond our unsupervised ReID complemented with better detectors in settings we tested.

\subsection{Experimental Setup}

% NEED TO REFINE MORE
\noindent{\bf Datasets}: We evaluate our performance on the standard multi-object tracking benchmark--  MOT Challenge -- which consists of several challenging pedestrian tracking sequences with frequent occlusions, crowded scenes with sequences varying in their angle of view, size of objects, camera motion, and frame rate. It contains two challenging tracking benchmarks, namely MOT16 and MOT17~\cite{milan2016mot16}. They both share the same training and testing sequences, but MOT16 provides only DPM~\cite{felzenszwalb2008discriminatively} detections, whereas MOT17 provides two additional sets of public detections (namely  Faster R-CNN~\cite{ren2015faster} and SDP~\cite{yang2016exploit}) and has more accurate ground truth. The primary metrics used for measuring performance are MOTA~\cite{bernardin2008evaluating} and IDF1, which are a combination of simpler metrics like False Positives, False Negatives, and ID Switches. 

\noindent{\bf Implementation details}: We obtain our SimpleReID model by training a ResNet50~\cite{he2016deep} backbone popularly used by trackers for a fair comparison. We train the model with tracklets generated by SORT~\cite{bewley2016simple} on the PathTrack~\cite{manen2017pathtrack} dataset to test generalization to unseen MOT16/17 data. We perform analysis studies on the entire training dataset and report results on MOT Challenge hidden test set~\footnote{The MOT Challenge web page: ~\url{https://motchallenge.net}.}. Our model was implemented using PyTorch and Torchreid~\cite{zhou2019torchreid} and trained on a GTX1080Ti GPU. For any tracker used \cite{wojke2017simple,bergmann2019tracking}, we utilize the implementations provided by the authors, leaving all the hyperparameters unchanged and simply replacing their supervised ReID model with SimpleReID. We use the CenterTrack model trained with single images w.r.t augmentations and incorporate the SimpleReID model using the DeepSORT framework. Our code and pretrained models will be released upon acceptance of the paper.

\begin{table}[t]
\begin{center}
\resizebox{0.935\textwidth}{!}{
\begin{tabular}{lllcccccc} \toprule
 \textbf{Detector}  & \textbf{Method}  & \textbf{Published}  & \textbf{Unsup}   & \textbf{MOTA$\uparrow$} & \textbf{IDF1$\uparrow$} & \textbf{IDSw$\downarrow$} & \textbf{FP$\downarrow$} & \textbf{FN$\downarrow$} \\ \midrule
\multicolumn{9}{c}{MOT16} \\ \hline
 \multirow{4}{*}{Batch} & GCRA~\cite{ma2018trajectory} & ICME18 & $\times$  & 48.2 & 48.6 & 821 & 5104 & 88586 \\
                        & HCC~\cite{ma2018customized} & ACCV18 & $\times$   & 49.3 & 50.7 & \textbf{391} & 5333 &86795 \\
                        & LMP~\cite{tang2017multiple} & CVPR17  & $\times$  &  48.8 & 51.3 & 481 & \textbf{6654} & 86245\\
                        & MPN~\cite{braso2019learning} & CVPR20 & $\times$  & \textbf{55.9} & \textbf{59.9} & 431 & 7086 & \textbf{72902}\\ \midrule
\multirow{4}{*}{Online} & AMIR~\cite{sadeghian2017tracking} & ICCV17 & $\times$ & 47.2 & 46.3 & 774 & 2681 & 92856 \\
                        & KCF~\cite{chu2019online}  & WACV19 & $\times$  &   48.8 &  47.2  & 906 & 5875 & 86567   \\
                        & RAR16~\cite{fang2018recurrent} & WACV18 & $\times$ & 45.9 & 48.8 & 648 & 6871 & 91173 \\
                       & MOTDT~\cite{chen2018real} & ICME18 & $\times$ & 47.6 & 50.9 & 792 & 9253 & 85431 \\
                       & STRN~\cite{xu2019spatial} & ICCV19 & $\times$ & 48.5 & 53.9 & 747 & 9038 & 84178 \\
                       & DeepMOT~\cite{xu2020train} & CVPR20 & $\times$ &  54.8  &  53.4  & 645 & 2955 & 78765  \\
                       & CenterTrack~\cite{zhou2020tracking} & Arxiv20* & \checkmark & 62.2 & 54.1 & 1677 & 5433 & \textbf{61767} \\
                       & DMAN~\cite{zhu2018online} & ECCV18 & $\times$ & 46.1 & 54.8 & \textbf{532} & 7909  & 89874 \\
                       & Tracktor++v2 ~\cite{bergmann2019tracking} & ICCV19 & $\times$ &  56.2    &  54.9  & 617 & \textbf{2394}  &  76844 \\
                       & MIFT & Arxiv20* & $\times$ & 60.1 & 56.9 & 739 & 6964  & 65044 \\
                       & \textbf{Ours} & - &  \checkmark & \textbf{62.4} & \textbf{58.5} & 588 & 5909 & 61981  \\ \midrule
                       \multicolumn{9}{c}{MOT17} \\ \hline
\multirow{4}{*}{Batch} & MHT~\cite{kim2015multiple} & CVPR15 & $\times$ &  50.7 & 47.2 & 6543 &46638  & 224955 \\
                       & FWT~\cite{henschel2017improvements} & CVPRW18 & $\times$   & 51.3 & 47.6 & 2648 &24101  & 247921 \\
                       & MHT-bLSTM~\cite{kim2018multi}& ECCV18 & $\times$ & 47.5 & 51.9 & 2069 & 25981 & 268042 \\
                       & jCC~\cite{keuper2018motion} & TPAMI18 & $\times$ & 51.2 & 54.5 & 1802 & 25937 & 247822 \\     
                       & MPN~\cite{braso2019learning} & CVPR20 & $\times$   & \textbf{55.7} & \textbf{59.1} & \textbf{1433} & \textbf{25013} & \textbf{223531}\\\midrule
\multirow{4}{*}{Online} & FAMNet~\cite{chu2019famnet} & ICCV19 & $\times$   &  52.0    & 48.7  & 3072 & 14138 & 253616    \\
                        & DeepMOT~\cite{xu2020train} & CVPR20 & $\times$ & 56.7 & 52.1 & 2351 & 8895 & 233206 \\ 
                        & MOTDT~\cite{chen2018real} & ICME18 & $\times$ & 50.9 & 52.7 & 2474 & 24069 & 250768 \\
                        & CenterTrack~\cite{zhou2020tracking} & Arxiv20* & \checkmark & 61.4 & 53.3 & 5326 & 15520 & \textbf{196886} \\  & Tracktor++v2~\cite{bergmann2019tracking} & ICCV19 & $\times$ & 56.3 & 55.1 & 1987 & \textbf{8666} & 235449 \\
                        & DMAN~\cite{zhu2018online} & ECCV18 & $\times$ & 48.2 & 55.7 & 2194 & 26218 & 263608 \\
                        & MIFT~\cite{huang2020refinements} & Arxiv20* & $\times$ & 60.1 & 56.4 & 2556 & 14966 & 206619 \\
                        & STRN~\cite{xu2019spatial} & ICCV19 & $\times$ & 50.9 & 56.5 & 2397& 25295 & 249365 \\
                        & \textbf{Ours} & - & \checkmark  & \textbf{61.7} & \textbf{58.1} & \textbf{1864} & 16872 & 197632 \\\bottomrule
\end{tabular}}
\end{center}
\caption{Results on the MOT Challenge test set benchmark using public detections. Unsup indicates approach does not need supervision (no tracking labels required). * are recent parallel works. Up/down arrows indicate higher/lower is better.}
\label{tab:test_set}
\end{table}
\subsection{MOT Challenge Benchmark Evaluation}

We submit our best performing unsupervised tracker to the MOT Challenge Benchmark. The submitted tracker consists of our proposed SimpleReID model incorporated with CenterTrack \cite{zhou2020tracking} for bounding box regression using public detections. We compare the performance on the MOT Challenge test set with state-of-the-art supervised trackers and provide results in Table \ref{tab:test_set}.  Surprisingly, we observe that our developed unsupervised tracker outperforms all supervised trackers on MOT16/17 setting a new state-of-the-art in terms of MOTA and IDF1 scores among all trackers on public detections. 

We beat the previous best tracker (CenterTrack) by 0.2/0.3 MOTA and 4.4/4.8 IDF1 scores on MOT16/MOT17, respectively. The significant increase in IDF1 score can be entirely attributed to the efficacy of our SimpleReID model, because while CenterTrack is a good detector, it cannot  maintain long tracks which is  compensated by using our appearance features for Re-identification. We reduce ID switches made by CenterTrack by nearly 3x, achieving the lowest ID switches compared to other online trackers. 

\subsection{Analysis}

Past literature \cite{tang2017multiple,ma2018customized} indicates that unsupervised ReID is unlikely to excel in crowded scenarios due to the complexities of tracking in such scenes. In this subsection, we provide two sets of evidence to demonstrate that SimpleReID indeed performs well across diverse scenarios: (i) We show that the test performance of SimpleReID (on unseen videos) is equivalent to that of a supervised ReID model, on its training set itself (ii) We show that SimpleReID achieves the above desiderata even with simple trackers which are highly reliant on the ReID component.
\begin{table}[t]
\begin{center}
\resizebox{0.7\textwidth}{!}{
\begin{tabular}{lcc|lcc} \toprule
 \textbf{ReID}       & \textbf{MOTA$\uparrow$} & \textbf{IDF1$\uparrow$} &   \textbf{ReID}    & \textbf{MOTA$\uparrow$} & \textbf{IDF1$\uparrow$} \\ \midrule
%\multicolumn{6}{c}{MOT15} \\ \hline
% \multicolumn{3}{c|}{ACF} & \multicolumn{3}{c}{HTC} \\ 
%Random     & 50.8 & 55.4  & Random     &      &      \\ 
%ImageNet   & 50.8 & 55.4  & ImageNet   & 56.4 & 58.4 \\
%Ours       & 52.0 & 59.2  & Ours       & 58.0   & 62.3 \\
%Supervised & 52.0 & 59.0  & Supervised & 58.0   & 62.1 \\ \midrule
\multicolumn{6}{c}{MOT16} \\ \hline
 \multicolumn{3}{c|}{DPM} & \multicolumn{3}{c}{POI} \\ 
None     &  57.6    & 62.0      & None     &  68.3    & 67.6     \\
ImageNet   & 57.6 & 62.0  & ImageNet   & 68.3      &    67.7  \\
Ours       & 57.6 & \textbf{62.6}  & Ours       &  68.5    & \textbf{69.5}     \\ 
Supervised & 57.6 & \textbf{62.5}  & Supervised &    68.5  &  \textbf{69.4}    \\ \midrule 
\multicolumn{6}{c}{MOT17} \\ \hline
\multicolumn{3}{c|}{FRCNN} &  \multicolumn{3}{c}{POI} \\ 
None     & 61.6 & 64.6 & None     &     68.5 & 67.6     \\
ImageNet   & 61.6 & 64.7 & ImageNet   & 68.5 & 67.6 \\
Ours       & 61.7 & \textbf{65.2} & Ours       & 68.6 & \textbf{69.4} \\
Supervised & 61.7 & \textbf{65.2} & Supervised & 68.6 & \textbf{69.3} \\ \bottomrule
\end{tabular}}
\end{center}
\caption{Ablation study comparing the performance of different ReID models within the Tracktor~\cite{bergmann2019tracking} framework. We observe that our unsupervised SimpleReID achieves the same performance (IDF1 scores) as supervised ReID. DPM, FRCNN and POI correspond to different detectors.}
\label{tab:traktor_compare}
\end{table}
\begin{table}[t]
\begin{center}
\resizebox{0.7\textwidth}{!}{
\begin{tabular}{lcc|lcc}\toprule

\textbf{ReID}       & \textbf{MOTA$\uparrow$} & \textbf{IDF1$\uparrow$} & \textbf{ReID}       & \textbf{MOTA$\uparrow$} & \textbf{IDF1$\uparrow$} \\ \hline
\multicolumn{3}{c}{MOT16-POI} & \multicolumn{3}{c}{MOT17-POI} \\ \midrule 
No ReID    & 58.1 & 57.1  & No ReID    & 57.9 & 56.9 \\
Random     & 51   & 34.6  & Random     & 50.7 & 34.3 \\
ImageNet   & 60.3 & 62   & ImageNet   & 59.9 & 61.6 \\
Market1501 & 60.3 & 61.5 & Market1501 & 59.9 & 61.1\\
Ours       & 60.5 & \textbf{65.9} & Ours       & 60.1 & \textbf{65.5} \\
Supervised & 60.4 & \textbf{65.9} & Supervised & 60   & \textbf{65.5} \\ \bottomrule
\end{tabular}}
\end{center}
\caption{Ablation study comparing the performance of different ReID models within the DeepSORT~\cite{wojke2017simple} framework. We observe that our unsupervised SimpleReID achieves the same performance (IDF1 scores) as supervised ReID.}
\label{tab:deepsort_compare}
\end{table}

\textbf{Limits of unsupervised ReID:} We test the limits of SimpleReID by comparing the performance of our model with supervised models. We perform experiments across various weaker scenarios such as having no ReID, or using pretrained-ImageNet as-is, and show that these perform significantly worse than SimpleReID - proving that SimpleReID is important to match supervised performance. We first train another recent supervised tracker, Tracktor++v2\cite{bergmann2019tracking}, which uses bounding box regression along with a supervised ReID model to predict the position of an object in the next frame. We train the supervised ReID model on the training data for MOT16/ MOT17 and then benchmark the performance on the same training set. In contrast, this data is new to our SimpleReID models, i.e., have not seen these videos previously. Our experiment results are tabulated in Table \ref{tab:traktor_compare}. We observe that using ImageNet-pretrained ReID somewhat improves IDF1 scores compared to using no ReID network at all, but fails to achieve the upper bound by a considerable margin. Our SimpleReID approach successfully recovers the remaining performance gap. This is achieved consistently across different variations.

\textbf{ReID-reliant unsupervised tracking:} Due to the low dependence of Tracktor on its ReID model, one may argue that it might not be the best framework for evaluation of ReID models in tracking. Hence, we also perform the same experiments on a popular tracker DeepSORT \cite{wojke2017simple} that is highly reliant on the ReID model used, since the only visual features it receives is from the ReID network. We replace the supervised ReID model used in DeepSORT with different ReID methods and tabulate results in Table \ref{tab:deepsort_compare}. First, we observe that replacing supervised ReID with random features causes a severe drop in performance over supervised counterpart, with MOTA score decreasing by 9.4\% and IDF1 decreasing by 31.3\%, demonstrating the degree of reliance on ReID in the DeepSORT framework. When substituted with features from an ImageNet-pretrained ResNet, we get a similar result: a significant improvement over SORT, yet much lower than supervised ReID performance. We further benchmark with a supervised ReID model trained on Market1501 dataset~\cite{zheng2015scalable} and observe lower performance compared to the ImageNet-pretrained model, indicating that features learned for cross-camera person-ReID datasets without trajectory annotations do not transfer to multi-object tracking. Lastly, we observe that our unsupervised SimpleReID covers the remaining performance gap, as seen above.

\begin{table}[t]
\begin{center}
\resizebox{0.83\textwidth}{!}{
    \begin{tabular}{ccc|cc|c}\toprule
       \textbf{Detector} & \multicolumn{2}{c}{\textbf{SimpleReID}} & \multicolumn{2}{c}{\textbf{Oracle ReID+Kill+MM}} & \\ \midrule
     & MOTA$\uparrow$ & IDF1$\uparrow$ & MOTA$\uparrow$ & IDF1$\uparrow$ & IDF1 Gain  \\ \hline
    %  & \multicolumn{2}{c}{No ReID} & \multicolumn{2}{c}{With SimpleReID} & \\ \hline
    %       YOLOv3~\cite{redmon2016you} & 55.9 & 61.5 & 56.5 & 62.5 & 1.0 \\
    %       DPM~\cite{felzenszwalb2008discriminatively} & 57.3 & 61.6 & 58.5 & 62.9 & 1.3 \\
    %       Faster-RCNN~\cite{ren2015faster} & 61.6 & 64.6 & 61.7 & 65.2 & 0.6 \\
    %       HTC~\cite{chen2019hybrid} & 66.9 & 66.9 & 67.7 &  68.1 & 1.2\\
    %       SDP~\cite{yang2016exploit} & 67.4 & 66.7 & 67.7 & 68.1  & 1.4 \\
    %       POI~\cite{yu2016poi} & 68.5 & 67.6 & 68.6 & 69.4 & 1.8 \\ \midrule

         YOLOv3~\cite{redmon2016you} & 56.5 & 62.5  & 61.5 & 66.1 & 3.6\\
         DPM~\cite{felzenszwalb2008discriminatively} & 58.5 & 62.9  & 62.4 & 66.3 & 3.4 \\
         Faster-RCNN~\cite{ren2015faster}  & 61.7 & 65.2  & 65.5 &  68.5 & 3.3 \\
         HTC~\cite{chen2019hybrid} & 67.7 &  68.1 & 75.6 & 70.5 & 2.4  \\
         SDP~\cite{yang2016exploit} & 67.7 & 68.1 & 73.0 & 70.6 & 2.5\\
         POI~\cite{yu2016poi} & 68.6 & 69.4 & 73.5 & 71.4 & \textbf{2.0} \\ 
         \bottomrule
    \end{tabular}}
\end{center}
\caption{Ablation study comparing the difference between performance of SimpleReID across detectors on MOT17. We observe that the difference decreases from 3.6 to 2.0 with improved detectors.}
\label{tab:oracle_compare}
\end{table}
\textbf{Scope for improvement in ReID:} We further explore the best performance achievable by a ReID network using the Tracktor framework and explore the scope for further improvement of our SimpleReID. To obtain the possible best performance, we test Tracktor with an Oracle ReID \cite{bergmann2019tracking} and observe that there is a 3.3 IDF1 score gap between SimpleReID and the Oracle. We repeat the same experiment with the latest off-the-shelf detectors and tabulate the results in Table \ref{tab:oracle_compare}. We observe that with modern detectors, the gap between SimpleReID and the corresponding oracles is small enough to limit the scope for further improvement.

Overall, we conclude that unsupervised SimpleReID counterintuitively matches the limiting performance of supervised counterparts in difficult MOT scenarios, by leveraging only unlabeled videos. Since our model works in extreme cases such as DeepSORT, where tracking is entirely reliant on the ReID model for encoding appearance information, we expect that the efficacy of SimpleReID will generalize to other trackers as well. We demonstrated the potential of unsupervised trackers by outperforming all supervised MOT16/17 trackers, setting a new state-of-the-art in MOTA and IDF1 scores and performing close to the optimal ReID. If it is indeed generalizable, we believe that this work has significant implications for research in supervised ReID for tracking.

\section{Conclusion}
\label{sec:conclusion}
 
We propose the first step in the direction of developing unsupervised re-identification for MOT and demonstrate that our simple approach performs at par with supervised counterparts across diverse setups. When combined with recent unsupervised association models \cite{xu2020train,bergmann2019tracking}, we obtain accurate unsupervised trackers. The tracker we submit ranks first in the MOT Challenge, beating all the latest supervised approaches. Our investigation suggests reconsideration on whether the shift towards more complex, supervised, end-to-end MOT models is necessary. We hope our work is useful to sidestep high annotation costs otherwise thought to be a requirement necessary to feed the data-hungry supervised trackers.

%\clearpage
\bibliographystyle{splncs04}
\bibliography{egbib}
\end{document}